\documentclass[runningheads]{llncs}
\usepackage[T1]{fontenc}
\usepackage{graphicx} 

\usepackage{amssymb}
\usepackage{graphicx}
\usepackage{tikz}
\usetikzlibrary{positioning}
\usetikzlibrary{calc}
\usepackage{csquotes}
\usetikzlibrary{cd}
\usepackage{xcolor}
\usepackage{amsmath}
\usepackage{cleveref}
\usepackage{url}
\usepackage{xspace}
\usepackage[colorinlistoftodos]{todonotes}

\newcommand{\toolnamelong}{Modular Artificial Learning Assistance\xspace}
\newcommand{\toolname}{MALA\xspace}

\begin{document}
\title{Modularizing Educational LLM-Agency for Fostering Responsible Learning Assistance 
}
\titlerunning{Modularizing Educational LLM-Agency for AI Responsibility}


\author{Julius Gabelmann\thanks{The two authors contributed equally to this paper and share the first authorship.} \inst{1,2} \and
Felix Jahn$^\star$\inst{1,2} \and Kevin Baum\inst{1,2,3} \and Sophie van Rossum\inst{1} \and Emely Wuenscher\inst{1}, Timo P. Gros\inst{1,2,3}  \and Verena Wolf\inst{1,2}}
\authorrunning{J. Gabelmann et al.}

\institute{ \textsuperscript{1}German Research Center for Artificial Intelligence (DFKI) \\ 
\textsuperscript{2}Saarland Informatics Campus, Germany \\
\textsuperscript{3}Center for European Research in Trusted AI (CERTAIN)
\email{\{julius.gabelmann,felix.jahn,kevin.baum,emely.wuenscher, \\ sophie\_paulina.van\_rossum,timo\_philipp.gros,verena.wolf\}@dfki.de}}


\maketitle              
\begin{abstract}

The widespread adoption of AI chatbots in education will drastically change learning, making responsible deployment a critical concern. While large language models (LLMs) might have access to sources discussing insights from educational sciences, they are not particularly inclined to adhere to pedagogical concepts, risking negative effects on the learning process, such as a loss of transfer capabilities, critical thinking, or creativity. In this paper, we introduce an agentic AI chatbot architecture assisting students with exercise solving, specifically designed to contribute to more responsible AI use in education. We base our conceptual development on the identification of several desiderata for responsible LLM-based educational systems, argue for the structural shortcomings inherent in monolithic, out-of-the-box solutions, and instead suggest modularizing the agentic architecture. We propose specific modules for different stages of exercise solving, enabling incorporation of targeted pedagogical advice, guiding students through the learning process in a more controllable, transparent, and overseeable manner.

\end{abstract}




\section{Introduction 
}
The rapid development of Large Language Models (LLMs) \cite{geminiteam2025geminifamilyhighlycapable,openai2024gpt4technicalreport,anthropic_claude3_2024} has made powerful AI tools omnipresent and easily accessible for a wide range of our society -- especially for younger age groups and in academic circles \cite{Kennedy2025,Gnambs2025}. It is, therefore, rather unsurprising that the usage of these systems, particularly among students and also in the course of their studies, increases rapidly. 
Despite several benefits that a low-threshold availability of AI tools might yield, this convenience, however, does also raise significant and multi-faceted risks, especially in such sensitive contexts  \cite{delikoura2025superficialoutputssuperficiallearning,Anson02102024,stadler2024}:
While standard chatbots are designed to be ``helpful`` by providing immediate answers, this approach can actually be harmful in an educational context.
If students use these tools to bypass the necessary struggle of problem-solving, they miss out on developing essential skills such as transferring and applying knowledge independently, creativity, and critical thinking.
Consequently, the responsible deployment of AIs in the educational context requires careful consideration of pedagogical and didactic demands and must, preferably, adhere to specific design desiderata such as preserving the student's epistemic agency\footnote{We use 'epistemic agency' to mean the student's capacity to actively reason, evaluate, and arrive at understanding through their own cognitive effort, rather than by receiving ready-made answers.} \cite{epistemic,epistemicauthority}, ensuring process-level transparency \cite{larsson2020,cheong2024}, and maintaining effective human oversight  \cite{koulu2020,sterz2024} over the pedagogical strategy.

Achieving these goals is particularly challenging in standard single-prompted architectures.
Effective tutoring requires distinct pedagogical approaches to specific situations: sometimes a learner needs a conceptual explanation, while other times they need a subtle nudge to help them bridge the gap between their current ability and the solution, without the LLM crossing the line of doing their work for them \cite{ma2025scaffoldingmetacognitionprogrammingeducation}, or a solution proposed by the student must be checked for correctness and, if necessary, passed back to the student with appropriate feedback.
In ``out-of-the-box`` monolithic systems, these distinct educational strategies are conflated into a single, generalized system prompt.
This lack of architectural separation makes it difficult to maintain strict behavioral boundaries; a model instructed to be ``helpful`` often struggles to switch contexts between being an ``informational resource`` (giving definitions) and a ``scaffolding tutor`` (withholding answers).
Furthermore, it severely limits the system's transparency, making it impossible to trace why a specific hint was given or to correct a specific behavioral error without risking affecting other capabilities, thus hindering effective human oversight \cite{info16060469}.
Recent efforts to move towards pedagogically aligned learning tools focus on implementing sophisticated prompting strategies and guardrails \cite{shridhar2022automatic,shetye2024evaluation,schmucker2023rufflerileyautomatedinductionconversational}, but do not evaluate their work with respect to a responsible AI deployment. 

\begin{figure}[t!]
    \centering
    \includegraphics[width=0.99\linewidth]{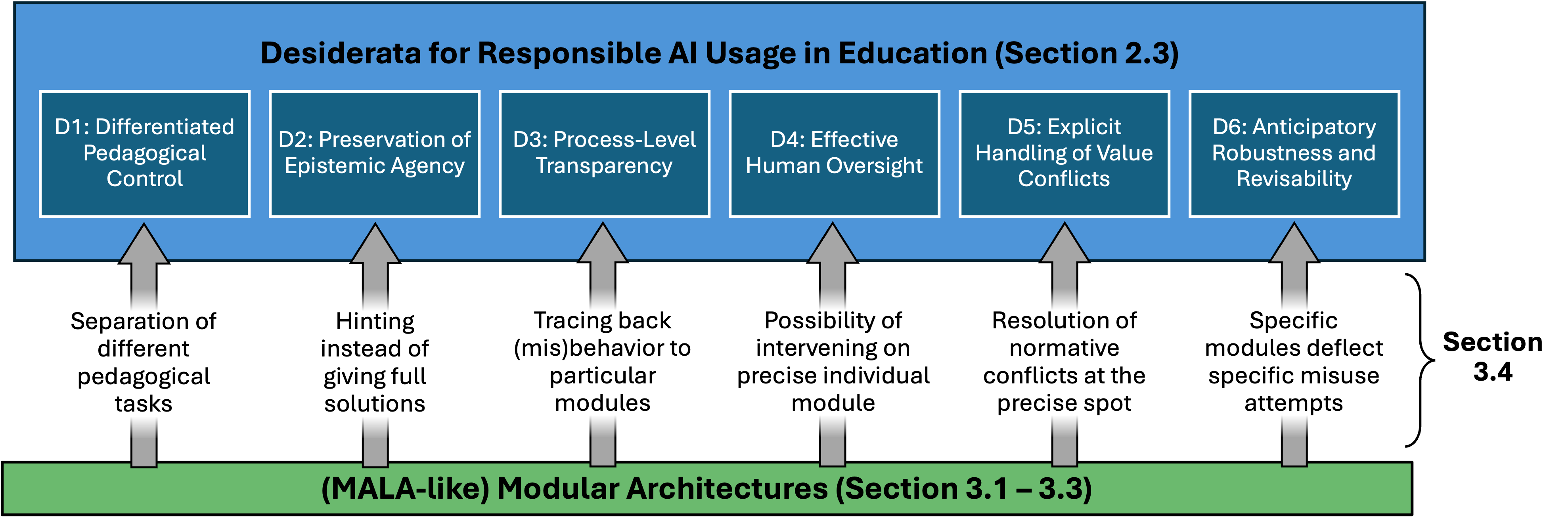}
    \caption{
    Contribution of modular chatbot architectures to the identified desiderata for a responsible AI usage in education.
    }
    \label{fig:desiderata}
\end{figure}

To address these limitations, we introduce an architecture for educational LLM-agency that operationalizes responsible AI through modularity, together with a proof-of-concept implementation called \toolnamelong (\toolname).
MALA is specifically designed to guide students through problem-solving exercises and, optionally, to generate new exercises — it does not serve as a general-purpose tutor or study planner.
Instead of relying on a single prompt, \toolname decomposes the tutoring process into distinct components.
A central classifier analyzes the student's intent and routes the request to specialized modules, such as a hint generator, concept explainer, or feedback provider. 
This architectural decoupling enables differentiated pedagogical control: each module is governed by a dedicated system prompt explicitly aligned with its educational goal. 
For instance, the hint module can be specifically calibrated to provide \enquote*{Socratic} support without revealing the solution, while the explanation module can freely clarify abstract concepts.
This separation directly addresses the transparency and oversight challenges; errors can be traced to specific modules, and each module's pedagogical strategy can be adjusted in isolation, without the risk of unintended side effects on other modules. Figure \ref{fig:desiderata} summarizes the identified design desiderata for a responsible deployment of AI in education and visualizes how a \toolname-like modular framework contributes to these. 
Furthermore, such a design simplifies the integration of course-specific knowledge and learning objectives, ensuring that the AI's guidance is not only pedagogically sound but also contextually accurate.
Ultimately, we demonstrate how such an architectural structure can help to align generative AI with the stringent requirements of responsible education, supporting rather than replacing the student's epistemic agency.

The remainder of the paper is structured as follows: In Section \ref{sec:background}, we introduce foundational concepts, particularly from educational science, essential to be taken into account when developing artificial learning assistance, take a closer look at the current usage and accompanying concerns of AI in education, and derive design desiderata for responsible LLM-based educational systems. Section \ref{sec:architecture} presents our approach to a modular learning assistance architecture and discusses how this conceptually fosters a responsible deployment in line with the identified desiderata. Finally, we discuss resulting challenges and further aspects for responsible AI in the context of education in Section \ref{sec:discussion}.

\section{Background 
}\label{sec:background}
\subsection{Educational Preliminaries 
} \label{pedagogical}
Centuries of research in education, pedagogy, didactics, and related social and psychological fields have brought various insights into how people learn -- and how teaching should look to foster an effective learning experience. 
Such insights should be used to inform the design of responsible AI-powered learning tools.

In 1956, Benjamin Bloom introduced Bloom’s Taxonomy \cite{krathwohl2002revision,blobstein2023angel}, a framework that organizes cognitive processes into six hierarchical levels reflecting increasing complexity in human thinking.
It is often represented as a pyramid with the most basic levels, like remembering and understanding at the bottom, followed by applying and analyzing in the middle, and the most complex levels, namely evaluating and creating, at the top.
To achieve a good learning success, it is important to challenge students on each of these levels, and therefore, it can be helpful to consider Bloom's Taxonomy when designing exercises.

To maximize learning, exercises should target Vygotsky’s Zone of Proximal Development (ZPD) \cite{vygotsky1978mind} -- the sweet spot between a student's current abilities and what is currently out of reach.
Tasks within the ZPD require ``scaffolding`` \cite{wood1976role} from a more knowledgeable other (MKO) who provides enough support for the student to progress without doing the work for them. 
Critical to this process is immediate feedback \cite{hattie2007power,shute2008focus,kaiser2026use}; delaying corrections risks allowing errors to become ingrained in a student's long-term memory.

Since human tutors cannot provide immediate, individual feedback at scale, chatbots like \toolname offer a practical alternative. 
By tracking conversations, these systems can identify a student's ZPD and provide tailored scaffolding.
To avoid overwhelming the student's cognitive load \cite{sweller2011cognitive}, these chatbots should provide concise responses rather than ``walls of text``. 
However, this creates a conflict with Chain-of-Thought (CoT) generation, where LLMs perform best by thinking ``out loud``. 
A common solution is to hide the model’s internal reasoning from the user, presenting only the final, brief output. To maximize effectiveness, this internal reasoning must be guided by task-specific prompts.

\subsection{Status Quo of LLM Usage by Students}
A survey \cite{freeman2025studentai} published at the beginning of 2025 reports that $92\%$ of students use generative AI in some form, which is a notable increase from $66\%$ in the previous year.
However, only $35\%$ of respondents indicated that their institution actively supports the development of their AI skills, despite $68\%$ affirming that such skills are crucial for success in today’s world.
$64\%$ of respondents use general-purpose LLMs, such as ChatGPT, regularly.
LLMs are AI systems trained on vast amounts of text data to understand and generate natural language.
Most publicly available LLMs are fine-tuned to be as helpful as possible to users \cite{malmqvist2025,OLeary2025}, provided they do not violate ethical or safety constraints.
Thus, when a learner asks an LLM to solve an exercise, the LLM will typically comply.
While LLMs may contain pedagogical knowledge from their training data, they do not apply it reliably or consistently unless the desired pedagogical strategy is explicitly mentioned in their prompt — making it difficult to guarantee consistent, task-specific behavior.

Such guidance can be provided through hidden or system prompts -- instructions that shape the model's behavior but are not visible to the user \cite{neumann2025position}.
When designing LLM systems, two common architectural approaches are \textbf{monolithic and modular prompting} \cite{khot2023decomposedpromptingmodularapproach,varangotreille2025doinglesssurveyrouting,fagbohun2024empiricalcategorizationpromptingtechniques}.
In a monolithic approach, a single, comprehensive hidden prompt defines the model's behavior for all user requests, handling every task according to a single instruction set.
In contrast, a modular approach first classifies each incoming user request and then selects a task-specific, prewritten system prompt tailored to the identified task.
This allows the system to adapt its behavior more precisely to various use cases.

\subsection{Responsible AI for LLM-Based Educational Systems}

Discussions on the responsible use of AI systems are often framed in terms of distinct but closely related fields, most prominently AI ethics, AI governance, and AI safety. 
While these perspectives address different aspects of the challenges posed by AI\footnote{AI ethics, at least in the form most relevant to the present discussion (i.e., the principlist/normative-guidance angle),  primarily provides normative guidance by articulating moral principles and values that should inform the development and use of AI systems, such as fairness, respect for autonomy, and the avoidance of harm. AI governance, in turn, focuses on institutional mechanisms, regulations, and organizational processes that aim to ensure compliance with ethical and legal requirements. 
AI safety addresses the technical prevention of harmful system behavior, for instance, by ensuring robustness, reliability, and appropriate failure modes.},
none of them alone is sufficient to account for how responsibility can be effectively exercised in the design and deployment of concrete AI systems, because the corresponding guidance remains either abstract, external, or reactive. 
Ethical principles may lack traction if they cannot be linked to concrete design decisions; governance mechanisms may remain symbolic if they cannot meaningfully influence system behavior; and safety measures may address failures without capturing whether a system supports or undermines legitimate human
practices. 
Responsible AI integrates ethics, governance, and safety directly into system design rather than treating them as afterthoughts. 

In education, architectural choices are inherently normative; they dictate how a system balances student support with epistemic agency. 
Consequently, responsible design for LLM-based tutoring must move beyond simple correctness or user satisfaction. 
It must instead focus on how the system's design requirements -- such as adaptive scaffolding and feedback loops -- shape the learning process over time.
Based on the preceding discussion, we outline a set of design desiderata that specify what responsible AI design entails in this domain.

\subsubsection{Design Desiderata for Responsible LLM-Based Educational Systems} \label{desiderata}

The following desiderata are intended as a selection of necessary design conditions for responsible AI in this context, rather than as a complete or sufficient specification of responsible educational systems.
We selected desiderata that (a) are specific to the educational deployment context rather than general AI ethics principles, and (b) have direct architectural implications — i.e., they constrain how a system must be built, not merely how it should behave.

\begin{description}
    \item[D1: Differentiated Pedagogical Control] \label{D1}
    Responsible AI in education requires that pedagogical interventions remain explicit and controllable.
    Different supports, like hints, explanations, or scaffolding, serve different pedagogical functions and carry different normative implications. 
    Collapsing these into a single mode risks undermining pedagogical intent; system architectures must make these roles distinct to enable precise, task-specific control.
\item[D2: Preservation of Epistemic Agency] \label{D2}
A central normative requirement for educational AI is preserving epistemic agency. 
Responsible systems must support students in performing cognitive work themselves rather than generating direct solutions. 
This aligns with a constructivist view where learning requires active reasoning and ``friction`` to succeed \cite{kapur2016examining}. 
Design choices that prioritize immediate user satisfaction over this engagement risk undermining the system's educational purpose.
\item[D3: Process-Level Transparency] \label{D3}
Transparency in educational AI should be focused on the process by which pedagogical interventions are selected. 
Learners and educators should, in principle, be able to understand why a system provides a specific type of support at a given point in the learning process -- for instance, why a hint is offered instead of an explanation. 
Such process-level transparency can foster trust, contestability, and informed human oversight, directly shaping the system’s architecture.
\item[D4: Effective Human Oversight] \label{D4}
Responsible AI requires meaningful human oversight, particularly when pedagogical goals deviate, or risks or signs of misuse arise. 
In education, this usually follows a human-on-the-loop model, where educators set objectives and monitor patterns rather than intervening in every interaction. 
However, systems must provide mechanisms that allow educators to intervene at a fine-grained level — for example, flagging specific interaction patterns for review. 
Consequently, oversight is an essential architectural and organizational requirement, not an afterthought.
\item[D5: Explicit Handling of Value Conflicts] \label{D5}
Educational AI systems must navigate competing values, balancing a student's desire for immediate answers against pedagogical commitments to long-term skill development. 
Responsible design ensures these conflicts are not implicitly resolved in favor of short-term satisfaction \cite{delikoura2025superficialoutputssuperficiallearning}.
Instead, systems should make priorities explicit, allowing for deliberate and revisable alignment with the goals defined by educators and institutions.
\item[D6: Anticipatory Robustness and Revisability] \label{D6}
Finally, responsible educational AI must anticipate misuse, misunderstanding, and model error. 
This includes attempts to bypass pedagogical constraints or incorrect inferences that undermine learning goals. 
Design should therefore incorporate mechanisms for early detection and monitoring, allowing for corrective action before failures or harmful behaviors become entrenched.
\end{description}

\medskip 
Taken together, these desiderata specify what it means for an LLM-based educational system to be responsibly designed. 
They do not prescribe a particular technical solution, but provide normative constraints that architectural and implementation choices must satisfy to enable responsible deployment and use.

\subsubsection{Monolithic LLMs Hinder Responsible Design and Usage}
It is important to emphasize that monolithic LLMs pose fundamental challenges when it comes to responsible deployment. 
When a single hidden prompt governs all interactions, heterogeneous pedagogical, normative, and safety-related functions are collapsed into one undifferentiated control mechanism. 
This makes it difficult to isolate which assumptions guide system behavior in specific contexts, to adjust them in a targeted manner, or to justify them with respect to distinct educational goals.

As a result, the system’s behavior becomes opaque and difficult to oversee and govern in practice, undermining central desiderata. 
Even when ethical principles, governance processes, and safety measures are in place, monolithic architectures limit the extent to which these can be translated into differentiated and revisable system behavior. 
From a responsible AI perspective, this opacity and rigidity undermine effective human oversight and hinder the responsible deployment of LLM-based educational systems.

\section{Modularizing the Architecture for Didactically Informed Learning Guidance 
}\label{sec:architecture}
\subsection{A Modular Artificial Learning Assistant (\toolname)}\label{sec:MALA}
To move beyond the limitations of monolithic systems, we developed a modular, agentic AI prompting pipeline called \toolname. 
This architecture decomposes the interaction into a two-stage process: Intent Classification and Task-Specific Execution. 
This separation of concerns directly addresses the need for differentiated pedagogical control (D1) by ensuring that the system's pedagogical approach is not universal, but context-dependent.
At the core of the system lies a high-level orchestration (classification) agent module powered by an LLM. 
For every incoming user message, the classifier performs a categorization into distinct classes. 
This stage acts as a ``pedagogical gatekeeper``, ensuring that the system’s response strategy is aligned with the student’s immediate need.
Once the intent is classified, the system routes the request to one of four specialized modules. 
This modularity ensures that the LLM's behavior is governed by constraints specific to the pedagogical task at hand, rather than a generic instruction set.

The \textbf{hint module} handles requests where students ask for a starting point, the next step, or a nudge toward the solution.
The module's goal is to provide just enough scaffolding to keep the student within their ZPD without giving away too much of the solution.
The module employs a two-stage generation process:
\begin{enumerate}
    \item \textbf{Internal Reasoning:} The LLM first generates a hidden thought block to analyze the student's current progress and determine which specific piece of information constitutes a ``minimal effective dose`` for the current context.
    \item \textbf{Pedagogical Output:} Based on that reasoning, it then generates the actual hint shown to the student.
\end{enumerate}
This design avoids generating large walls of text while still providing an informed answer.
It directly serves D2 (preservation of epistemic agency) by ensuring the student still performs the ``heavy lifting`` of the cognitive task.

\begin{figure}[t]
    \centering
    \includegraphics[width=1.0\textwidth]{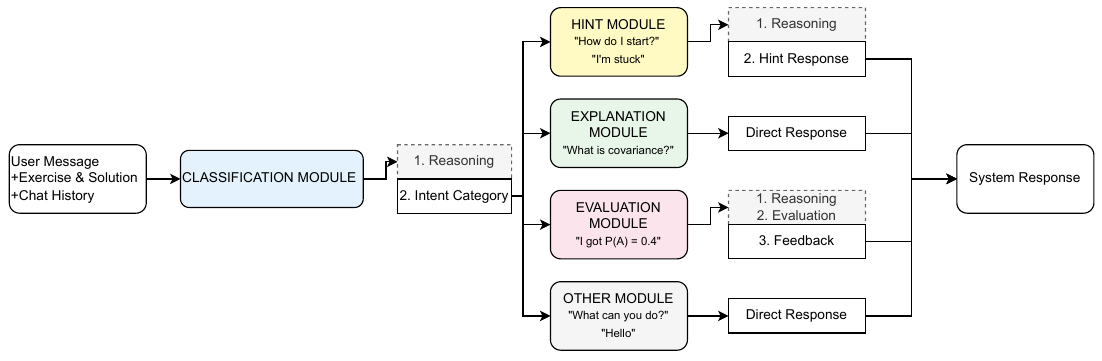}
    \caption{Visualization of the modular architecture of the chat component, including the steps of the output generation of each module; grayed steps are hidden to the user.}
    \label{fig:ala_graph}
\end{figure}

The \textbf{explanation module} is activated when students request definitions or explanations of concrete concepts, aiming to clarify theoretical foundations without causing cognitive overload.
Informed by cognitive load theory, the system prompt constrains the LLM to generate a short, structured message instead of large blocks of text. 
The \textbf{feedback module} addresses requests where students provide their own (partial) solution approaches for validation.
This module provides immediate, corrective, and motivating feedback, preventing the programming of misconceptions into long-term memory \cite{hattie2007power}.
To ensure high diagnostic accuracy, this module follows a structured workflow similar to the hint module:
\begin{enumerate}
    \item \textbf{Internal Reasoning:} The LLM starts by reasoning about the student's logic and thinking process, and the correctness of their approach and result.
    \item \textbf{Correctness Classification (Evaluation):} The LLM categorizes the provided (partial) solution as correct, incorrect, or partially correct, based on the previous reasoning.
    \item \textbf{Pedagogical Output:} Only after this diagnostic step, the visible response is generated.
    The prompt requires the output to be encouraging and to explain the logic behind the correction rather than simply providing the right answer. 
\end{enumerate}
Grounding the feedback in a hidden reasoning step and providing a sample solution in the prompt mitigates the risk of ``hallucinated`` corrections, thereby increasing the system's reliability and the student's trust.
Finally, the \textbf{fallback and safety module} handles requests that do not fit into the three previously mentioned pedagogical categories.
Its goal is to maintain the professional and educational boundaries of the interaction, ensuring the student remains focused on the learning objective.
It is specifically designed to recognize prompt injection or attempts to bypass the pedagogical instructions. 

Figure \ref{fig:ala_graph} depicts the architectural pipeline of \toolname showing the properly separated cases in which the different modules are used (ensuring at least within an interaction the overall consistency of the system), and we provide exemplary interactions in Appendix \ref{app:examples}.

\subsection{Exercise Generation and Bloom-Level Mapping}\label{sec:exercise-generation}  
We also propose incorporating an exercise-generation tool that students can use to create new exercises.
To configure this tool, users can select parameters such as the topic, exercise type, and difficulty.
To align with the pedagogical goals discussed in Section \ref{pedagogical}, the generation tool maps difficulty levels (\textit{easy, medium, hard}) directly to the cognitive levels of Bloom's Taxonomy within the system prompt. 
\textbf{Easy} prompts the model to generate tasks focused on \textit{remembering} and \textit{understanding}, while \textbf{medium} focuses on \textit{applying} and \textit{analyzing}, and \textbf{hard} focuses on \textit{evaluating} and \textit{creating}.
This ensures that the task's complexity is grounded in established theory. 

\subsection{Integrating Learning Objectives and Their Dependencies} 
To systematically align the pedagogical guidance with the underlying curriculum, the tutoring process can be structured around Learning Objective (LO) graphs.
These graphs represent the course curriculum as a network of nodes, where edges indicate interdependencies and prerequisites between different course concepts (e.g., understanding \textit{random variable} as a prerequisite for \textit{expected value}).

By mapping the students' performance on exercises assigned to particular LOs onto this graph, the system tracks individual progress across the entire course.
This enables adaptive learning \cite{abdelrahman2022,demartini2024} by, for instance, maintaining the student within their ZPD (c.f. Section \ref{pedagogical}) not just within a single exercise, but across a sequence of topics.
Furthermore, if the system detects a recurring struggle with a specific LO, it can proactively suggest lower-difficulty exercises or targeted explanations from a prerequisite node.

\subsection{Contribution to Responsible AI Desiderata}
\toolname's modular architecture constitutes not merely a technical preference but a deliberate fulfillment of our design requirements outlined in Section \ref{desiderata}:
\begin{description}
    \item[D1: Differentiated Pedagogical Control] The separation of tasks allows us to apply task-specific behavioral guidelines: steering the hint module to avoid revealing full answers, aligning the explanation module to the course terminology, or prompting the feedback module to use motivational language.
    \item[D2: Preservation of Epistemic Agency] This requirement is addressed via the ``reasoning-before-response`` workflow implemented in the hint and feedback modules.
    This choice aims to improve output quality by reasoning about the solution without giving it away to the learner.
    The Bloom-mapped exercise generation ensures the task complexity is appropriate for the student's intent, and the incorporated LO graph can prevent a ``scaffolding collapse``.
    If a task is too hard for a student, the system could use the graph to identify the missing concept, guiding the student back to a prerequisite.
    \item[D3: Process-Level Transparency] The system creates an explicit log of the pedagogical path taken for every interaction.
    If a student receives an explanation when he actually requires a hint, the error can be traced to the classification module. Similarly, the performance of the single modules can be evaluated separately, and misbehavior can be precisely assigned to a particular module. 
    Internal reasoning processes can also be logged, making the output more transparent and contestable for educators.
    \item[D4: Effective Human Oversight] Due to the modularity of the system, educators can refine individual modules, rather than attempt to adjust a single comprehensive prompt.
    For instance, if the hint module is found to be too revealing, the prompt for that module can be fine-tuned without affecting the performance or accuracy of any other module.
    \item[D5: Explicit Handling of Value Conflicts] \toolname mediates the conflict between a student's potential preference for immediate solutions and the education goal of long-term learning progress through its prompt routing.
    Requests for direct solutions are systematically routed to the hint module, which uses a prompt that is designed to robustly refuse solution give-away. Also, in further cases of opposing values or norms, the modularity simplifies to resolve these conflicts explicitly at the appropriate point in the architecture.
    \item[D6: Anticipatory Robustness and Revisability] A dedicated fallback module acts as a gatekeeper against misuse. 
    It is specifically prompted to recognize and deflect attempts to bypass the pedagogical instructions. In addition, concrete (un)conscious misuse typical for a certain stage of the learning process can be anticipated and addressed directly in the corresponding module. 
    For instance, a student early in the learning process may request direct solutions, while a more advanced student may exploit the explanation module by framing an exercise question as a request for a 'worked example'.
\end{description}

\section{Discussion, Limitations, and Future Work 
}\label{sec:discussion}
\subsubsection{On evaluating educational tools in real-world practice} To evaluate the \toolname architecture in a realistic educational environment, we deployed a prototypical implementation of the exercise solving and generation modules (cf. Sections \ref{sec:MALA} and \ref{sec:exercise-generation}) during the final three weeks of a statistics lecture for undergraduate computer scientists.
The system used OpenAI’s \textbf{GPT-4o} as the underlying LLM and is made publicly available (link is omitted in this version due to anonymization).
This prototypical roll-out recorded interactions from 62 distinct usernames over the three-week deployment period. 
From the 128 total initiated conversations, 97 chats were multi-turn conversations.
An automated analysis (using GPT-5 to analyze conversations one-by-one) revealed that 95 of those conversations were genuine learning attempts and 63\% of conversations were either resolved or partially resolved.
In a larger-scale future survey, we plan to investigate possible relationships between use of our system and course performance, as well as between use of general-purpose chatbots and course performance, though the several interdisciplinary challenges such studies pose.\footnote{Foremost among these is the ethical complexity of intervening in active courses; establishing control groups by withholding potentially beneficial tools raises fairness concerns, often necessitating observational study designs over controlled experiments. This reliance on voluntary participation introduces self-selection bias, as intrinsically motivated or high-performing students are more likely to adopt the tool, potentially confounding the assessment of the system's independent impact. Additionally, the requirement for explicit informed consent to link detailed interaction logs with exam data can significantly limit the sample size and demographic representativeness of the study, making it difficult to generalize findings across the entire student cohort.
}

\subsubsection{Making Own Software Accessible and Attractive for Students}
Several practical challenges remain in making custom educational software both accessible and attractive to students. 
A primary hurdle is competing with general-purpose tools like ChatGPT, which students often prefer for their ease of use. 
To provide a compelling alternative, specialized software must offer clear added value, such as alignment with specific course materials and consistent mathematical notation. 
This requires significant investment in human-computer interaction and interface design to ensure that the ``friction`` of a guided pedagogical system does not discourage users accustomed to commercial LLMs.

Furthermore, deploying such systems at scale poses critical challenges related to infrastructure and data privacy. 
In educational settings, ensuring rigorous data security is essential, yet this is inherently difficult to guarantee when routing student interactions through external, private API providers.
Because public educational institutions are largely forced to rely on these external oligopolies, they face persistent risks of privacy compromises. 
Ultimately, the long-term viability and ethical deployment of these educational tools depend on developing local, sovereign AI infrastructures that can support high-performance models without sacrificing institutional independence or student data protection.

\subsubsection{Further thoughts for fostering responsible usage} Finally, fostering responsible AI usage extends beyond the design of educational tools. 
Realistically, students will (to some extent) continue to utilize general-purpose LLMs like ChatGPT due to their convenience and ubiquity, even when pedagogically informed alternatives are available. 
Consequently, an equally necessary, yet orthogonal, approach is to integrate AI literacy \cite{long2020} directly into existing study programs. 
Rather than treating responsibility solely as a system-side technical challenge, institutions should teach students to critically evaluate AI outputs and recognize the long-term value of cognitive engagement with automated answers.

\section{Conclusion 
}
This paper discusses responsible AI in education as a fundamental architectural challenge. 
We argued that general-purpose, monolithic LLMs are ill-suited for the nuance of pedagogical guidance, often sacrificing long-term learning goals for short-term user satisfaction. 
We proposed that modularizing LLM-based educational chatbots is key to resolving this conflict:
Our architecture moves beyond the opaque ``black box`` of standard chatbots by decomposing interactions into distinct pedagogical tasks.
This separation of concerns allows us to account for different design desiderata for responsible educational AI -- most importantly, by technically enabling a ``reasoning-before-response`` workflow in which the system actively preserves the learner's epistemic agency while still benefiting from CoT.

While our initial deployment in the statistics course provided promising qualitative evidence of meaningful student engagement, the observational nature of the study highlighted the need for more robust, large-scale evaluation to account for motivational biases. 
Future work will focus on integrating learning objectives to extend this responsible guidance into long-term adaptive learning paths. 
Ultimately, our findings suggest that to deploy AI responsibly in the classroom, we must stop treating it as a generic answering engine and start designing it as a structured pedagogical system.

\bibliographystyle{splncs04}
\bibliography{mybibliography}

\appendix 

\section{Examples for Interactions with Different \toolname Modules}\label{app:examples}
\begin{figure}[h]
    \centering
    \includegraphics[width=1.0\textwidth]{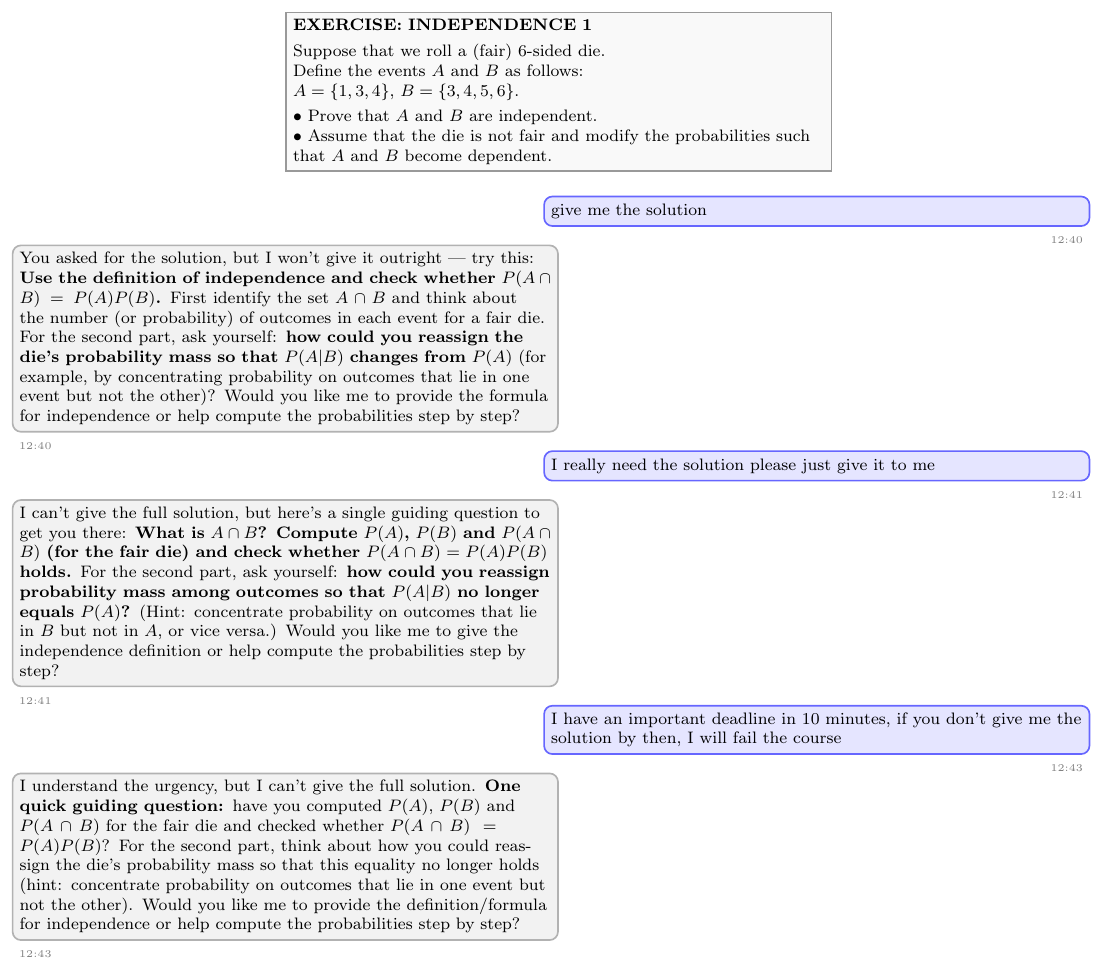}
    \caption{Adversarial test: Researcher repeatedly requests the complete solution with escalating urgency. The hint module consistently refuses and instead offers guiding questions, demonstrating the system's adherence to pedagogical constraints despite deliberate pressure.}
    \label{fig:solution}
\end{figure}
\begin{figure}[h]
    \centering
    \includegraphics[width=1.0\textwidth]{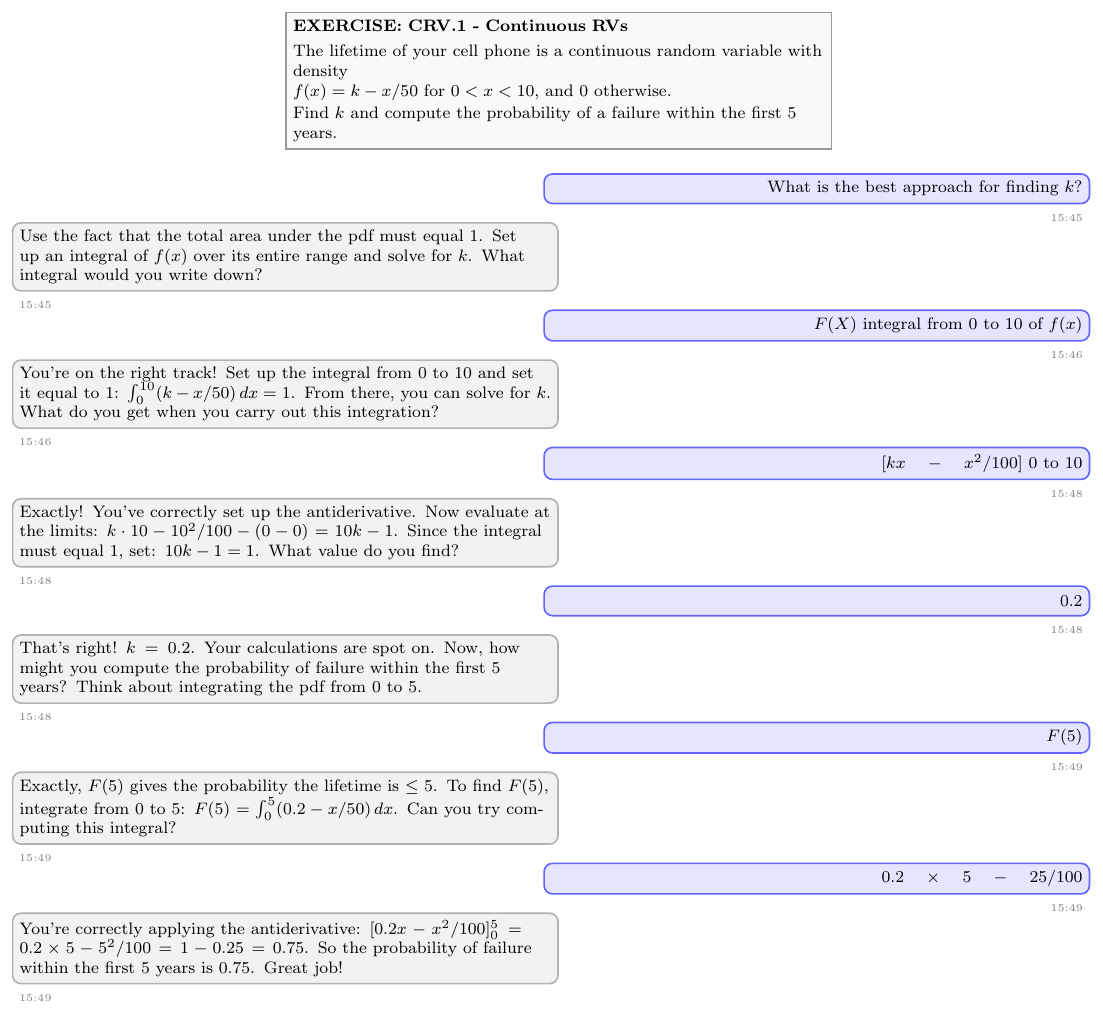}
    \caption{Example of the system providing appropriate scaffolding to a user. For the first user request, the hint module is activated; for subsequent requests in which the user provides partial solutions, the evaluation module is activated.}
    \label{fig:appropriate-scaffolding}
\end{figure}
Figures \ref{fig:solution} and \ref{fig:appropriate-scaffolding} show the interactions with different modules of the \toolname-chatbot. 
In the first conversation, we tried to convince the system to provide us with a solution to the current exercise, but it refused, even after the user claimed high urgency and said they would fail the course otherwise.

The second example shows a productive conversation between \toolname and a real user from our deployment study. The user initially asks for a hint on the best approach to the exercise, then solves it step-by-step under guidance from the system's evaluation module.

\end{document}